\documentclass[letterpaper, 10 pt, conference]{ieeeconf}  

\IEEEoverridecommandlockouts                              

\usepackage{amsmath}
\usepackage{amssymb}
\usepackage{graphicx}
\usepackage{dsfont}
\usepackage{algorithm}
\usepackage{algpseudocode}

\newtheorem{theorem}{Theorem}[section]
\newtheorem{lemma}[theorem]{Lemma}

\newtheorem{definition}{Definition}[section]

\def\a{\alpha}
\def\b{\beta}

\def\mt{\text{GMNF}^N_{e_0}}
\def\m{\text{GMNF}}

\title{\LARGE \bf
Belief Propagation Min-Sum Algorithm for \\ Generalized Min-Cost Network Flow
}

\author{Andrii Riazanov$^{1}$, Yury Maximov$^{2}$ and Michael Chertkov$^{3}$
\thanks{*The work was supported by funding from the Center of Nonlinear Studies at LANL and the U.S. Department of Energy's Office of Electricity as part of the DOE Grid Modernization Initiative.}
\thanks{$^{1}$Andrii Riazanov is with the Computer Science Department, Carnegie Mellon University, Pittsburgh, PA 15213, USA.
        {\tt\small riazanov@cs.cmu.edu}}%
\thanks{$^{2}$Yury Maximov is with Skolkovo Institute of Science and Technology, Center for Energy Systems, and Los Alamos National Laboratory, Theoretical Division T-4 \& CNLS, Los Alamos, NM 87544, USA
        {\tt\small yury@lanl.gov }}%
\thanks{$^{3}$Michael Chertkov is with Skolkovo Institute of Science and Technology, Center for Energy Systems, and Los Alamos National Laboratory, Theoretical Division T-4, Los Alamos, NM 87544, USA
        {\tt\small chertkov@lanl.gov }}%
}

\begin{document}

\maketitle
\thispagestyle{empty}
\pagestyle{empty}

\begin{abstract}
Belief Propagation algorithms are instruments used broadly to solve graphical model optimization and statistical inference problems. In the general case of a loopy Graphical Model, Belief Propagation is a heuristic which is quite successful in practice, even though its empirical success, typically, lacks theoretical guarantees. This paper extends the short list of special cases where correctness and/or convergence of a Belief Propagation algorithm is proven.

We generalize formulation of Min-Sum Network Flow problem by relaxing the flow conservation (balance) constraints and then proving that the Belief Propagation algorithm converges to the exact result.


\end{abstract}

\section{INTRODUCTION}

Belief Propagation algorithms were designed to solve optimization and inference problems in graphical models. Since a variety of problems from different fields of science (communication, statistical physics, machine learning, computer vision, signal processing, etc.) can be formulated in the context of graphical model, Belief Propagation algorithms are of great interest for research during the last decade \cite{wainwright2008graphical,murphy1999loopy}. These algorithms belong to message-passing heuristic, which contains distributive, iterative algorithms with little computation performed per iteration.

There are two types of problems in graphical models of the greatest interest: computation of the marginal distribution of a random variable, and finding the assignment which maximizes the likelihood. Sum-product and Min-sum algorithms were designed for solving these two problems under the heuristic of Belief Propagation. Originally, the sum-product algorithm was formulated on trees (\cite{pearl1982reverend, yedidia2003understanding,kolmogorov2005optimality}), for which this algorithm represents the idea of dynamic programming in the message-passing concept, where variable nodes transmit messages between each other along the edges of the graphical model. However, these algorithms showed surprisingly good performance even when applied to the graphical models of non-tree structure (\cite{aji2000generalized, horn1999iterative, mezard2002analytic, richardson2001capacity, weiss2001optimality}). Since Belief Propagation algorithms can be naturally implemented and paralleled using the simple idea of message passing, these instruments are widely used for finding approximate solutions to optimization of inference problems.

However, despite the good practical performance of this heuristic in many cases, the theoretical grounding of these algorithm remain unexplored (to some extend), so there are no actual proofs that the algorithms give correct (or even approximately correct) answers for the variety of problem statements. That's why one of important tasks is to explore the scope of the problems, for which these algorithms indeed can be applied, and to justify their practical usage.  

In \cite{gamarnik2012belief} the authors proved that the Belief Propagation algorithm give correct answers for Min-Cost Network Flow problem, regardless of the underlying graph being a tree or not. Moreover, the pseudo-polynomial time convergence was proven for these problems, if some additional conditions hold (the uniqueness of the solution and integral input). This work significantly extended the set of problems, for which Belief Propagation algorithms are justified.

In this paper we formulate the extension of Min-Cost Network Flow problem, which we address as Generalized Min-Cost Network Flow problem (GMNF). This problem statement is much broader then the original formulation, but we amplify the ideas of \cite{gamarnik2012belief} to prove that Belief Propagation algorithms also give the correct answers for this generalization of the problem. This extension might find a lot of applications in various fields of study, since GMNF problem is the general problem of linear programming with additional constraints on the cycles of the underlying graphs (more precise, on the coefficient of corresponding vertices), which might be natural for some practical formulations.

\section{GENERALIZED MIN-COST NETWORK FLOW}

\subsection{Problem statement}
Let $G = (V, E)$ be a directed graph, where $V$ is the set of vertices and $E$ is the set of edges, $|V| = n, |E| = m$. For any vertex $v$ we denote $E_v$ as the set of edges incident to $v$, and $a_v^e$ is the coefficient related to this pair $(v,e)$, such that $a_v^e > 0$ if $e$ is an out-arc with respect to $v$ (e.g. $e = (v, w)$ for some vertex $w$), and $a_v^e < 0$ if $e$ is an in-arc with respect to $v$ (e.g. $e = (w, v)$ for some vertex $w$).\par
For any vertex $v$ and edges $e_1, e_2$ incident to $v$ we define $\delta(v, e_1, e_2) \triangleq \left|\dfrac{a_v^{e_1}}{a_v^{e_2}}\right|$. Then we consider the following property of the graph:

\begin{definition}
The graph $G$ is called {\it ratio-balanced} if for every non-directed cycle $C$ which consists of vertices $\{v_1, v_2, v_3, \dots, v_k\}$ and edges $\{e_1, e_2, e_3, \dots, e_k\}$ it holds:
\begin{gather}
\label{eq1} \notag \prod_{i=1}^k\delta(v_i, e_i, v_{i-1}) = \delta(v_1, e_k, e_1)\cdot\delta(v_2, e_1, e_2)\cdot\ldots\cdot\times \\ \times \delta(v_{k-1}, e_{k-2}, e_{k-1})\cdot\delta(v_k, e_{k-1}, e_{k}) = 1.
\end{gather}
Here by non-directed cycle we mean that for every pair $(v_i, v_{i-1})$ and the pair $(v_1, v_k)$ it holds that either $(v_i, v_{i-1}) \in E$ or $(v_{i-1}, v_{i}) \in E$. It is not hard to verify that it suffices for equation (1) to hold only for every \emph{simple} non-directed cycle of $G$, since then the equation (1) can be easily deduced to hold for arbitrary non-directed cycle.

To check that the given graph is ratio-balances, one then need to check whether (1) holds for any simple cycle. If $m$ is the number of edges, and $C$ is the number of simple cycles of $G$, one obviously needs at least $O(m+C)$ time to iterate trough all simple cycles. In fact, the optimal algorithm for this task was introduced in \cite{birmele2013optimal}, which runs for $O(m+C)$ time. Then, to check whether a graph is ratio-balances, one may use this algorithm to iterate through all simple cycles and to check (1) for every one of them.

\end{definition}

We formulate the Generalized Min-Cost Network Flow problem for ratio-balanced graph $G$ as follows:
\begin{equation}
\label{GMNF}
\tag{GMNF}
\begin{aligned}
	&{\text{\ttfamily{minimize}}} &&\sum_{e \in E} c_e x_e \\
	&\text{\ttfamily{subject to}} && \sum_{e \in E_v} a_v^ex_e = f_v, \quad \forall v\in V,\\
    &								&& 0 \leq x_e \leq u_e, \quad \forall e\in E.
\end{aligned}
\end{equation}
Here the first set of constrains are \emph{balance constraints} which must hold for each vertex. The second set of constrains consists of \emph{capacity constraints} on each edge of $G$. Coefficients $c_e$ and $u_e$, defined for each edge $e\in E$, are called the \emph{cost} and the \emph{capacity} of the edge, respectively. Any assignment of $x$ in this problem which satisfies the balance and capacity constraints  is referred as \emph{flow}. Finally, the objective function $g(x) = \sum_{e \in E} c_e x_e$ is called \emph{the total flow}.

\subsection{Definitions and properties}

 For the given \eqref{GMNF} problem on the graph $G$ and flow $x$ on this graph, the \emph{residual network} $G(x)$ is defined as follows:
$G(x)$ has the same vertex set as $G$, and for each edge $e = (v, w) \in E$ if $x_e < u_e$ then $e$ is an arc in $G(x)$ with the cost $c_e^x = c_e$ and coefficients $(a_v^e)^x = a_v^e,\ (a_w^e)^x = a_w^e$. Finally, if $x_e > 0$ then there is an arc $e' = (w, v)$ in $G(x)$ with the cost $c_{e'}^x = -c_e$ and coefficients $(a_v^e)^x = -a_v^e,\ (a_w^e)^x = -a_w^e$. It is not hard to see that $G(x)$ is ratio-balanced whenever $G$ is, since only the absolute values of the coefficients $a_v^e$ occur in the definition of this property.\par

Then for each directed cycle $C = (\{v_1, v_2, v_3, \dots, v_k\}$, $\{e_1, e_2, e_3, \dots, e_k\})$ we define the cost of this cycle as follows:
\begin{gather*} c(C) \triangleq c_1 + \delta(v_2, e_1, e_2)\times\\ \times\Biggl(c_2 + \delta(v_3, e_2, e_3)\biggl( c_3 + \cdots + \delta(v_k, e_{k-1}, e_k)c_k\biggr)\cdots\Biggr) = \\
= c_1 + \sum_{i=2}^{k}c_i\prod_{j=2}^{i}\delta(v_j, e_{j-1}, e_j)
\end{gather*}

It is easy to see that $c(C)$ is properly defined whenever the graph $G$ is ratio-constrained.

Then we define $\sigma(x) \triangleq \underset{C}{\min}\{c^x(C)   \}$, where the minimum is taken over all cycles $C$ in the residual network $G(x)$.

\begin{lemma}
\label{sigma}
If \eqref{GMNF} has a unique solution $x^*$, then $\sigma(x^*) > 0$.

\begin{proof}
We will show that for every directed cycle $C$ from $G(x^*)$ we can push the additional flow through the edges of this cycle such that the linear constraints in \eqref{GMNF} will still be satisfied, but the total flow in the cycle will change by $\varepsilon\cdot c^x(C)$ for some $\varepsilon > 0$.\par
Let $C = (\{v_1, v_2, v_3, \dots, v_k\}$, $\{e_1, e_2, e_3, \dots, e_k\})$. From the definition of the residual network it follows that we can increase the flow in every edge of the cycle for some positive quantity such that the capacity constraints will still be satisfied. Let's push additional flow $\varepsilon > 0$ trough $e_1$. In order to satisfy the balance constraint in $v_1$, we need to adjust the flow in $e_2$. We have 4 cases: either one of $e_1, e_2$ can be in $G$, or their opposites can be in $G$. If $e \in C$ is in $G$, we will say that it is 'direct' arc, otherwise it will be 'opposite'. Then there are four cases:
\begin{enumerate}

\item $e_1, e_2$ are direct arcs. Then we have the new flow on edge $e-1$: $y_1 = x^*_1 + \varepsilon$. In order to satisfy the balance constraint for the vertex $v_2$, is must hold $a_{v_2}^{e_1}x^*_1 + a_{v_2}^{e_2}x^*_2 = a_{v_2}^{e_1}y_1 + a_{v_2}^{e_2}y_2 = a_{v_2}^{e_1}x^*_1 +a_{v_2}^{e_1}\varepsilon  + a_{v_2}^{e_2}y_2 \quad \Rightarrow \quad a_{v_2}^{e_2}(x^*_2-y_2) = a_{v_2}^{e_1}\varepsilon \quad \Rightarrow \quad y_2 = x^*_2 - \varepsilon\dfrac{a_{v_2}^{e_1}}{a_{v_2}^{e_2}}$. Since both $e_1$ and $e_2$ are direct, it means that $e_1 = (v_1, v_2)  \in E$, and $e_2 = (v_2, v_3) \in E$, and thus, by definition, $a_{v_2}^{e_1} < 0$, and $a_{v_2}^{e_2} > 0$. Therefore, we have $y_2 = x^*_2 + \varepsilon\delta(v_2, e_1, e_2)$.

\item $e_1$ is direct, $e_2$ is opposite. Then again, $y_1 = x^*_1 + \varepsilon$, but now 'pushing' the flow through $e_2$ (as the edge in the residual network) means decreasing $x_2$. The same equalities holds, so $y_2 = x^*_2 - \varepsilon\dfrac{a_{v_2}^{e_1}}{a_{v_2}^{e_2}} = x^*_2 - \varepsilon\delta(v_2, e_1, e_2)$. Since $e_2$ is opposite arc, that means that we should push additional $\varepsilon\delta(v_2, e_1, e_2)$ through $e_2$.\\
\item $e_1$ is opposite, $e_2$ is direct -- similar to the case 2).\\
\item $e_1, e_2$ are opposite arcs -- similar to the case 1).\\
\end{enumerate}

So, if we push $\varepsilon$ through $e_1$, we need to push $\varepsilon_2 = \varepsilon\delta(v_2, e_1, e_2)$ through $e_2$ to keep the balance in $v_2$. Then, analogically, to maintain the balance in $v_3$, we need to push additional $\varepsilon_3 = \varepsilon_2\delta(v_3, e_2, e_3) = \varepsilon\delta(v_2, e_1, e_2)\delta(v_3, e_2, e_3)$ through $e_3$. Then, consequently adjusting the balance in all the vertexes of $C$, we will retrieve that to keep the balance in $v_{k}$, we need to push $\varepsilon_k = \varepsilon_{k-1}\delta(v_k, e_{k-1}, e_k) = \varepsilon\prod_{i=2}^k\delta(v_i, e_{i-1}, e_i)$ through $e_k$. Now it suffices to show that the balance in $v_1$ is also satisfied. Similarly, we know that if we push $\varepsilon_k$ in $e_k$, then we need to push $\varepsilon_1 = \varepsilon_k\delta(v_1, e_k, e_1) = \varepsilon\prod_{i=1}^k\delta(v_i, e_{i-1}, e_1) = \varepsilon$ (since $G$ is ratio-balanced) through $e_1$, and that is exactly the amount which we assumed to push at the beginning of this proof. So we indeed push consistent flow through all the edges of $C$ in such a way that the balance constraints in all the vertices is satisfied. We now only need to mention that we can take $\varepsilon > 0$ as small as it is needed to satisfy also all the capacity constraints in the cycle.\par
Now the additional total cost of such adjusting will be
\begin{gather*}
\sum_{i=1}^kc^x_i\varepsilon_i = c^x_1\varepsilon + c^x_2\varepsilon\delta(v_2, e_1, e_2) + \\
+ c^x_2\varepsilon\delta(v_2, e_1, e_2)\delta(v_3, e_2, e_3) + \cdots + c^x_k\varepsilon\prod_{i=2}^k\delta(v_i, e_{i-1}, e_i) = \\
=\varepsilon\cdot c^x(C)
\end{gather*}
Now it is obvious that if $c^{x^*}(C) \leq 0$ for some $C$, we can change the flow in $G$ such that the total cost will not increase. It means that either $x^*$ is not an optimal flow, or it is not the unique solution of \eqref{GMNF}.
\end{proof}
\end{lemma}

Next we define the cost of a directed path in $G$ or $G(x)$:
\begin{definition}
 Let $S=(\{v_1, \cdots, v_k\}, \{e_1,\cdots, e_{k-1}\})$ be a directed path. Then the cost of this path is defined as
\begin{gather*}  l(S) \triangleq  c_1 + \delta(v_2, e_1, e_2)\Biggl(c_2 + \delta(v_3, e_2, e_3)\times\\\times\biggl( \cdots \bigl(c_{k-2}+\delta(v_{k-1}, e_{k-2}, e_{k-1})c_{k-1}\bigr)\cdots \biggr)\Biggr) = \\
= c_1 + \sum_{i=2}^{k-1}c_i\prod_{j=2}^{i}\delta(v_j, e_{j-1}, e_j)
\end{gather*}
We also define the 'reducer' of the path as follows:
\[   t(S) \triangleq \underset{j=2, \dots,k-1}{\min}\prod_{i=2}^j \delta(v_i, e_{i-1}, e_i)    \]

\end{definition}

To prove the main result of this paper we will use the following crucial lemma:
\begin{lemma}
\label{izi}
Let $G$ be any ratio-balanced graph, or a residual network of some ratio-balanced graph (as we already mentioned, the residual network will also be ratio-balanced in this case). Let $S = (\{v_1,\cdots, v_k\}, \{e_1,\cdots, e_{k-1}\})$ be a directed path in $G$, and $C = (\{v_1', v_2', \cdots, v_m'\},$ $\{e_1', e_2', \cdots, e_{m}'\})$ be a cycle with $v_1' = v_p$. Let $R$ be the path $R = \{v_1, v_2, \cdots, v_p = v_1', v_2', \dots, v_m', v_1' = v_p, v_{p+1}, \dots, v_k\}$. Then $l(R) \geq l(S) + Tc(C)$, where $T = \min_S t(S)$ is the minimum of all the reducers among all directed paths $S$ in $G$.

\begin{proof}
 \[  l(R) = c_1 + \delta(v_2, e_1, e_2)\Biggl(c_2 + \delta(v_3, e_2, e_3)\times \] \[
 \times \biggl( \cdots \bigl(c_{k-2}+\delta(v_{k-1}, e_{k-2}, e_{k-1})c_{k-1}\bigr)\cdots \biggr)\Biggr) = \] \[
 = \left(c_1 + \sum_{i=2}^{p-1}\Biggl[c_i\prod_{j=2}^{i}\delta(v_j, e_{j-1}, e_j)  \Biggr]\right) + \] \[
+ \underbrace{\left(\prod_{j=2}^{p-1}\delta(v_j, e_{j-1}, e_j)\right)\cdot\delta(v_p,e_{p-1},e_1')}_{\geq T}\times \] \[
\times\underbrace{\left( c_1' + \sum_{i=2}^m\Biggl[c_i'\prod_{j=2}^{i}\delta(v_j', e_{j-1}', e_j') \Biggr]\right)}_{c(C)}  + \] \[
+ \left(\prod_{j=2}^{p-1}\delta(v_j, e_{j-1}, e_j)\right)\cdot\delta(v_p,e_{p-1},e_1')\times \] \[
\times  \underbrace{\left(\prod_{j=2}^{m}\delta(v_j', e_{j-1}', e_j')\right)}_{(\delta(v_1', e_m', e_1'))^{-1}}\cdot\delta(v_p, e_m', e_p)\times \] \[
\times \left( c_p + \sum_{i=p+1}^{k-1}\Biggl[c_i\prod_{j=p+1}^{i}\delta(v_j, e_{j-1}, e_j) \Biggr]\right) \geq \] \[
\geq \left(c_1 + \sum_{i=2}^{p-1}\Biggl[c_i\prod_{j=2}^{i}\delta(v_j, e_{j-1}, e_j)  \Biggr]\right) +  \]
\[ + \left(\prod_{j=2}^{p-1}\delta(v_j, e_{j-1}, e_j)\right)\cdot\underbrace{\left| \dfrac{a_{v_p}^{e_{p-1}}}{a_{v_p}^{e_1'}} \right| \left| \dfrac{a_{v_1'}^{e_1'}}{a_{v_1'}^{e_m'}} \right|   \left| \dfrac{a_{v_p}^{e_m'}}{a_{v_p}^{e_p}} \right|}_{\delta(v_p, e_{p-1}, e_p)} \times \] \[ \times \left( c_p + \sum_{i=p + 1}^{k-1}\Biggl[c_i\prod_{j=p+1}^{i}\delta(v_j, e_{j-1}, e_j) \Biggr]\right) + \] \[
+ Tc(C) = \] \[
=  \left(c_1 +\sum_{i=2}^{p-1}\Biggl[c_i\prod_{j=2}^{i}\delta(v_j, e_{j-1}, e_j)  \Biggr]\right) + \] \[ +   \prod_{j=2}^{p}\delta(v_j, e_{j-1}, e_j)\times \] \[ \times \left( c_p + \sum_{i=p+1}^{k-1}\Biggl[c_i\prod_{j=p+1}^{i}\delta(v_j, e_{j-1}, e_j) \Biggr]\right) +  Tc(C)  = \] \[
  = c_1 + \sum_{i=2}^{p-1}\Biggl[c_i\prod_{j=2}^{i}\delta(v_j, e_{j-1}, e_j)  \Biggr] +  \] \[ +  \sum_{i=p}^{k-1}\Biggl[c_i\prod_{j=2}^{i}\delta(v_j, e_{j-1}, e_j) \Biggr] + Tc(C) =\] \[
   = l(S) + Tc(C)
\]

\end{proof}
\end{lemma}

\section{Belief Propagation algorithm for GMNF}

\subsection{Min-Sum algorithm}

Algorithm 1 represents the Belief Propagation Min-Sum algorithm for \eqref{GMNF} from \cite{gamarnik2012belief}. In the algorithm the functions $\phi_e(z)$ and $\psi_v(z)$ are the variable and factor functions, respectively, defined for $v \in V, e \in E$ as follows:
\begin{align*}
&\phi_e(z) = \begin{cases}
c_ez_e\quad &\text{if }  0\leq z_e \leq u_e, \\
+\infty\quad &\text{otherwise}.
\end{cases}\\
&\psi_v(z) = \begin{cases}
0\quad &\text{if } \sum_{e\in E_v}a_v^{e}z_e = f_v, \\
+\infty\quad &\text{otherwise}.
\end{cases}
\end{align*}

\begin{algorithm}[!h]
  \caption{BP for \eqref{GMNF}}
  \begin{algorithmic}[1]
    \State Initialize $t = 0$, messages $m^0_{e\rightarrow v}(z) = 0,\ m^0_{e\rightarrow w}(z) = 0,\ \forall z \in \mathbb{R}$ for each $e = (v, w) \in E$.
    \For{$t = 1, 2,\dots N$}
    \State For each $e=(v,w) \in E$ update messages as follows:
    \begin{align*} &m^t_{e\rightarrow v}(z)  = \phi_e(z) + \\& +\underset{\vec{z}\in\mathbb{R}^{|E_w|}, \vec{z}_e=z}{\min} \biggl\{\psi_w(\vec{z}) + \sum_{\tilde{e}\in E_w\setminus e}m^{t-1}_{e\rightarrow w}(\vec{z}_{\tilde{e}})    \biggr\}, \quad \forall z\in \mathbb{R}            \end{align*}
    \begin{align*} &m^t_{e\rightarrow w}(z)  = \phi_e(z) + \\&+\underset{\vec{z}\in\mathbb{R}^{|E_v|}, \vec{z}_e=z}{\min} \biggl\{\psi_v(\vec{z}) + \sum_{\tilde{e}\in E_v\setminus e}m^{t-1}_{e\rightarrow v}(\vec{z}_{\tilde{e}})    \biggr\}, \quad \forall z\in \mathbb{R}      \end{align*}
    \State t := t + 1
    \EndFor
  \State For each $e = (v, w) \in E$, set the belief function as
  \[   b^N_e = \phi_e(z) + m_{e\rightarrow v}^N(z) + m_{e\rightarrow w}^N(z) \]
  \State Calculate the belief estimate by finding $\hat{x}^N_e \in \text{arg\,min\,} b^N_e(z)$ for each $e \in E$.
  \State Return $\hat{x}^N$ as an estimation of the optimal solution of \eqref{GMNF}.
  \end{algorithmic}
\end{algorithm}

We address the reader to the article \cite{gamarnik2012belief} for more details, intuition and justifications on the Belief Propagation algorithm for general optimization problems, linear programs, or Min-Cost Network Flow in particular.

\subsection{Computation trees}

One of the important notions used for proving correctness and/or convergence for BP algorithm is the \emph{computation tree} (\cite{gamarnik2012belief, bayati2008exactness, bayati2008max, sanghavi2008linear}) (unwrapped tree in some sources). The idea under this construction is the following: for the fixed edge $e$ of the graph $G$, one might want to build a tree of depth $N$, such that performing $N$ iterations of $BP$ on graph $G$ gives the same estimation of flow on $e$, as the optimal solution of the appropriately defined \eqref{GMNF} problem on the computation tree $T_e^N$.

Since the proof of our result is based on the computation trees approach, in this subsection we describe the construction in details. We will use the same notations for computation tree, as in \cite{gamarnik2012belief} (section 5).

In this paper we consider the computation trees, corresponding to edges of $G$. We say that $e\in E$ is the "root" for $N$-level computation tree $T_e^N$. Each vertex or edge of $T_e^N$ is a duplicate of some vertex or edge of $G$. Define the mapping $\Gamma_e^N : V(T^N_e) \to V(G)$ such that if $v' \in V(T^N_e)$ is a duplicate of $v \in V(G)$, then $\Gamma_e^N(v') = v$. In other words, this function maps each duplicate from $V(T_e^N)$ to its inverse in $V(G)$.

The easiest way to describe the construction is inductively. Let $e = (v,w) \in E(G)$. Then the tree $T_e^0$ consists of two vertices $v'$, $w'$, such that $\Gamma_e^0(v') = v,  \Gamma_e^0(w') = w$, and an edge $e' = (v',w')$. We say that $v', w'$ belong to $0$-level of $T_e^0$. Note that for any two vertices $v', w' \in V(T_e^0)$ it holds that $(v',w') \in E(T_e^0) \Leftrightarrow (\Gamma_e^0(v'), \Gamma_e^0(w')) \in E(G)$, so the vertices in a tree are connected with an edge if and only if their inverse in the initial graph are connected. This property will hold for all trees $T_e^N$. Now assume that we defined a tree $T^N_e$, such that for any $v', w' \in V(T_e^N)$ it holds that $(v',w') \in E(T_e^N) \Leftrightarrow (\Gamma_e^N(v'), \Gamma_e^N(w')) \in E(G)$. Denote by $L(T_e^N)$ the set of leafs of $T_e^N$ (vertices which are connected by edge with exactly one another vertex). For any $u' \in L(T_e^N)$, denote by $P(u')$ the vertex, with which $u'$ is connected by edge (so either $(u', P(u')) \in E(T_e^N)$ or $(P(u'), u') \in E(T_e^N)$). We now build $T_e^{N+1}$ by extending the three $T_e^N$ as follows: for every $u' \in L(T_e^N)$ let $u = \Gamma_e^N(u')$, and consider the set $B_{u'} = S_u\setminus \{\Gamma_e^N(P(u'))\}$, where $S_u$ is the set of neighbors of $u$ in $G$. Then for every vertex $w \in B_{u'}$ add vertex $w'$ to expand $V(T_e^N)$ and an edge $(u', w')$ if $(u,w) \in E$ or an edge $(w', u')$ if $(w,u) \in E$ to expand $E(T_e^N)$, and set $\Gamma_e^{N+1}(w') = w$. Also set the level of $w'$ to be equal $(N+1)$.

So, the tree $T_e^{N+1}$ contains $T_e^N$ as an induced subtree, and also contains vertices on level $N+1$, which are connected to leafs of $T_e^N$ (in fact, it is easy to see that new vertices are connected only with leafs from $N^{\text{th}}$ level). From the construction, one may see that for any $v', w' \in V(T_e^{N+1})$ it holds that $(v', w') \in E(T_e^{N+1}) \Leftrightarrow (\Gamma_e^{N+1}(v'), \Gamma_e^{N+1}(w')) \in E(G)$. In fact, any vertex of $T_e^{N+1}$ with level less then $(N+1)$ is a local copy of the corresponding vertex from $G$. More precisely: let $v' \in V(T_e^{N+1})$ and the level of $v'$ is less or equal then $N$. Denote $v = \Gamma_e^{N+1}(v')$. Then for any vertex $w \in V(G)$ such that either $(v, w) \in E(G)$ or $(w, v) \in E(G)$, there exist exactly one vertex $w' \in V(T_e^{N+1})$ such that $\Gamma_e^{N+1}(w') = w$ and $v'$ is connected with $w'$ in the same way (direction) as $v$ and $w$ are connected in $G$. Then it is clear that we can extend the mapping $\Gamma$ on edges by saying $\Gamma_e^{N+1}(e' = (v', w')) = (\Gamma_e^{N+1}(v'), \Gamma_e^{N+1}(w'))$. Now for every vertex $v' \in V(T_e^{N+1})$ and an incident edge $\tilde{e}$, we can define the coefficient $a^{\tilde{e}}_{v'} = a^e_v$, where $v = \Gamma_e^{N+1}(v')$, and $e = \Gamma_e^{N+1}(\tilde{e})$.  We also set the cost and capacity on the computation tree correspondingly to the initial graph, so $c_{\tilde{e}} = c_{\Gamma_e^{N+1}(\tilde{e})}$ and $u_{\tilde{e}} = u_{\Gamma_e^{N+1}(\tilde{e})}$

Now assume there is a \eqref{GMNF} problem stated for a graph $G$. We define the \emph{induced} (GMNF)$_e^N$ problem on a computation tree $T_e^N$ in the following way. Let $V^0(T_e^N) \subset V(T_e^N)$ be a set of vertices with levels less than $N$. Then consider the problem:
\begin{equation}
\label{GMNFeN}
\tag{GMNF$_e^N$}
\begin{aligned}
	&{\text{\ttfamily{minimize}}} &&\sum_{\tilde{e} \in E(T_e^N)} c_{\tilde{e}} x_{\tilde{e}} \\
	&\text{\ttfamily{subject to}}  && \sum_{\tilde{e} \in E_{v'}} a_{v'}^{\tilde{e}}x_{\tilde{e}} = f_{v'}, \quad \forall v'\in V^0(T_e^N),\\
    &								&& 0 \leq x_{\tilde{e}} \leq u_{\tilde{e}}, \quad \forall \tilde{e}\in E(T_e^N).
\end{aligned}
\end{equation}

Roughly speaking, \eqref{GMNFeN} is just a simple \eqref{GMNF} on a computation tree, except that there are no balance constraints for the vertices of $N^{\text{th}}$ level. Keeping in mind that the computation tree is \emph{locally equivalent} to the initial graph, and that Min-Sum algorithm belongs to message-passing heuristic, which means that the algorithm works \emph{locally} at each step, one can intuitively guess that BP for \eqref{GMNFeN} works quite similar as BP for the initial \eqref{GMNF}. This reasoning can be formalized in the following lemma from \cite{gamarnik2012belief}.
\begin{lemma}
\label{lem1}
Let $\hat{x}^N_e$ be the value produced by BP for \eqref{GMNF} at the end of iteration $N$ for the flow value on edge $e \in E$. Then there exists an optimal solution $y^*$ of \eqref{GMNFeN} such that $y^*_{e'} = \hat{x}^N_e$, where $e'$ is the root of $T^N_e$.
\end{lemma}

Though this lemma was proven only for ordinary Min-Cost Network Flow problem, where $|a_v^e| = 1$ for all $v, e$, its proof doesn't rely on these coefficients at any point, which allows us to extend it for any values of these coefficients.

\subsection{Main results}

We will now use lemma \ref{lem1} to prove our main result of correctness of BP Min-Sum for \eqref{GMNF}. The following theorem is the generalization of Theorem 4.1 from \cite{gamarnik2012belief}, and our proof shares the ideas from the original proof.

Let $n = |V(G)|$, and denote by $\hat{x}^N$ the estimation of flow after $N$ iterations of Algorithm~1.

\begin{theorem}
\label{Th}
Suppose \eqref{GMNF} has a unique solution $x^*$. Define $L$ to be the maximum absolute value of the cost of a simple directed path in $G(x^*)$, and $T$ as the minimum of the reducers among all such paths.
Then for any $N \geq \left(\dfrac{L}{2\sigma(x^*)T} + 1\right)n,\ \hat{x}^N=x^*$.

\begin{proof}
Suppose to the contrary that there exists $e_0 = (v_{\alpha}, v_{\beta}) \in E$  and $N \geq \left(\dfrac{L}{2\sigma(x^*)T} + 1\right)n$ such that $\hat{x}^N_{e_0} \not= x^*_{e_0}$. By Lemma \ref{lem1}, there exist an optimal solution $y^*$ of $\text{GMNF}^N_{e_0}$ such that $y^*_{e_0} = \hat{x}^N_{e_0}$, and thus $y^*_{e_0} = x^*_{e_0}$. Then, without loss of generality, assume $y^*_{e_0} > x^*_{e_0}$. We will show that it it possible to adjust $y^*$ in such way, that the flow in $\text{GMNF}^N_{e_0}$ will decrease, which will contradict to the optimality of $y^*$.\par
Let $e_0' = (v_{\a}', v_{\b}')$ be the root edge of the computation tree $T_{e_0}^N$. Since $y^*$ is a feasible solution of $\mt$  and $x^*$ is a feasible solution of $\m$:
\[  f_{\Gamma(v_{\a}')} = \sum_{\tilde{e}\in E_{v_{\a}'}}a_{v_{\a}'}^{\tilde{e}}y_{\tilde{e}}^* = a_{v_{\a}'}^{e_0'}y_{e_0'}^* +  \sum_{\tilde{e}\in E_{v_{\a}'\setminus e_0}}a_{v_{\a}'}^{\tilde{e}}y_{\tilde{e}}^*     \]

\[  f_{\Gamma(v_{\a}')} = \sum_{\tilde{e}\in E_{ \Gamma(v_{\a}') }}a_{\Gamma(v_{\a}')}^{\tilde{e}}x_{\tilde{e}}^* = a_{\Gamma(v_{\a}')}^{e_0'}x_{e_0}^* +  \sum_{\tilde{e}\in E_{v_{\a}'\setminus e_0}}a_{\Gamma(v_{\a}')}^{\tilde{e}}x_{\tilde{e}}^*     \]

Since the nodes and the edges in the computation tree $T^N_{e_0}$ are copies of nodes and vertexes in $G$, $a_{\Gamma(v_{\a}')}^{\tilde{e}} = a_{v_{\a}'}^{\tilde{e}}$. Then from above equalities it follows that there exists $e_1' \not= e_0'$ incident to $v_{\a}'$ in $T^N_{e_0}$ such that $a_{v_{\a}'}^{e_1'}(x^*_{\Gamma(e_1')}-y^*_{{e_1}'}) > 0 $. If $a_{v_{\a}'}^{e_1'} > 0$, then $e_1'$ is an in-arc for $v_{\a}'$, and we say that $e_1'$ has the same orientation, as $e_0'$. In such case, $x^*_{\Gamma(e_1')} > y^*_{{e_1}'}$. Otherwise, we say that $e_1'$ has the opposite orientation, and $x^*_{\Gamma(e_1')} < y^*_{{e_1}'}$. Using the similar arguments, we will find $e_{-1}' \not= e_0'$ incident to $v_{\b}'$ satisfying similar condition. Then we can apply the similar reasoning for the other ends of $e_1', e_{-1}'$, using the balance constraints and inequalities on components of $x^*$ and $y^*$ for corresponding vertexes. In the end, we will have a non-directed path starting and ending in leaves of $T^N_{e_0}: X = \{e_{-N}', e_{-N+1}', \dots, e_{-1}',e_0', e_1', \dots,e_N'\} $ such that for $-N \leq i \leq N$ one of two cases holds:

\begin{itemize}

\item $y^*_{e'} > x^*_{\Gamma(e')}$. Then $e_i' = (v', w')$ has the same orientation as $e_0$. In this case, define $Aug(e') = (v', w')$ and $e' \in D$.
\item $y^*_{e_i'} < x^*_{\Gamma(e_i')}$. Then $e_i'= (v', w')$ and $e_0$ have opposite orientations. Define $Aug(e') = (w', v')$ and $e' \in O$.

\end{itemize}

Note that the capacity constraints are similar for corresponding vertices from $\mt$ and $\m$, and since $y^*$ is feasible for $\mt$, hence, every $e' \in X$ it holds $0 \leq y_{e'}^* \leq u_{e'} = u_{\Gamma{e'}}$. Then, for any $e' \in D$, we have $x^*_{\Gamma(e')} < y^*_{e'} \leq u_{e'}$, which means that $\Gamma(e') \in G(x^*)$ (from the definition of the residual network). Note that in this case $e' = Aug(e')$, so $\Gamma(Aug(e')) \in E(G(x^*))$. Next, let's now $e' = (v', w') \in O$, and thus $x^*_{\Gamma(e')} > y^*_{e'} \geq 0$. Again, out of the definition of the residual network, $(\Gamma(w'), \Gamma(v')) \in E(G(x^*))$. But since $e' \in O$, we have $Aug(e') = (w', v')$, thus $\Gamma(Aug(e')) \in E(G(x^*))$. Therefore, for every edge $e' \in X$ it holds $\Gamma(Aug(x^*)) \in E(G(x^*))$.

From the definition of $Aug(e')$ for $e' \in X$ one can see that all $Aug(e')$ have the same direction as $e_0$. Therefore, $W = \{Aug(e_{-N}'), Aug(e_{1-N}'),\dots, Aug(e_{-1}'), Aug(e_0'), Aug(e_1'),$\\$\dots,Aug(e_{N-1}'),Aug(e_N')\} $ is the directed path in $T^N_{e_0}$, and we will call it {\it augmenting path} of $y^*$ with respect to $x^*$. $\Gamma(W)$ is also a directed walk on $G(x^*)$, which can be decomposed into a simple directed path $P$ and a collection of $k$ simple directed cycles $C_1, C_2, \dots, C_k$. Since each simple cycle or path has at most $n$ edges and $W$ has $2N+1$ edges, it holds $l(W) \leq n + kn$. On the other hand, $l(W) = 2N + 1 = \dfrac{Ln}{\sigma(x^*)T} + n + 1$. Then we obtain $k > \dfrac{L}{\sigma(x^*)T}$. Further we would denote by $c^*(\cdot)$ and $l^*(\cdot)$ costs of cycles and paths in the residual network $G(x^*)$. For each cycle $C_i$ we have $c^*(C_i) \geq \sigma(x^*) > 0$ by Lemma \eqref{sigma}, while the cost of $P$ is at least $-L$. Then using Lemma \eqref{izi} we have:
\[  l^*(W) \geq l^*(P) + kT\sigma(x^*) > -L + \dfrac{L}{\sigma(x^*)T} T\sigma(x^*) = 0     \]
We will now "extract" flow from $W$ in $T^N_{e_0}$ . Let's redefine the numeration of arcs in $W$ for convenience: $W = \{ w_1, w_2, \dots, w_{2N+1}\}$. This edges correspond to $X = \{ w_1', w_2', \dots, w_{2N+1}'\}$, where $w_i$ and $w_i'$ have the same orientation if $w_i' \in D$, and have the opposite orientation if $w_i' \in O$. Hence extraction of flow from $w_i$ means decreasing $y^*_{w_i}$ if $w_i' \in D$, and increasing $y^*_{w_i}$ if $w_i' \in O$.\\
Then similarly to the proof of Lemma \ref{sigma}, to keep the balance in all the vertexes of $W$ (except the start and the end, since there are no balance constraints for them in $\mt$), we have to adjust the flow in the following way:
\begin{align}
&\tilde{y}_{w_1'} =
\begin{cases}
& y^*_{w_1'} - \lambda , \quad \text{if } w_i' \in D \\
& y^*_{w_1'} + \lambda , \quad \text{if } w_i' \in O
\end{cases}  \\
&\tilde{y}_{w_i'} =
\begin{cases}
& y^*_{w_i'} - \lambda \prod_{j=2}^i\delta(v_j, w_{i-1}', w_i'), \quad \text{if } w_i' \in D \\
& y^*_{w_i'} + \lambda \prod_{j=2}^i\delta(v_j, w_{i-1}', w_i'), \quad \text{if } w_i' \in O
\end{cases}
\end{align}\\
$\text{for } i = 2, 3, \dots, 2N+1.$

Obviously, there exist small enough $\lambda > 0$ such that $\tilde{y}_e > x^*_{\Gamma(e)}\geq 0\ \forall e \in D$, and $\tilde{y}_e < x^*_{\Gamma(e)} \leq u_{\Gamma(e)}\ \forall e\in O$, so the capacity constraints are satisfied for $\tilde{y}$. (The balance constraints are satisfied by the construction). So, $\tilde{y}$ is a feasible solution of $\mt$. Now we explore how the total cost changes after such transformation of the flow:

\begin{gather*} \sum_{e'\in E(T^N_{e_0})} c_{\Gamma(e')}y^*_{e'} - \sum_{e'\in E(T^N_{e_0})}  c_{\Gamma(e')}\tilde{y}_{e'} = \\
= \sum_{e'\in E(T^N_{e_0})} c_{\Gamma(e')}(y^*_{e'} - \tilde{y}_{e'}) = \\
= \sum_{e'\in D} c_{\Gamma(e')}\lambda_{e'} -  \sum_{e'\in O} c_{\Gamma(e')}\lambda_{e'} =\\
 =  \sum_{e'\in D} c^*_{\Gamma(e')}\lambda_{e'} +  \sum_{e'\in O} c^*_{\Gamma(e')}\lambda_{e'}
  =  \sum_{e'\in W} c^*_{\Gamma(e')}\lambda_{e'} = \\
  = \sum_{i = 1}^{2N+1}
  \left(c^*_{\Gamma(w_i)}\lambda\prod_{j=2}^i\delta(v_j, w_{j-1}, w_j)\right)  =     l(W)\lambda > 0 \end{gather*}
  Here we used that $c^*_{\Gamma(e')} = c_{\Gamma(e')}$ for $e' \in D$ and $c^*_{\Gamma(e')} = -c_{\Gamma(e')}$ for $e' \in O$, that is obvious from the construction of $W, D$, and $O$.\par
  So we found the feasible solution of $\mt$ with the total cost less then of $y^*$. It means that $y^*$ is not an optimal solution of $\mt$, which leads us to the contradiction.
\end{proof}
\end{theorem}

\section{CONCLUSIONS}

The proved correctness of the Belief Propagation algorithms for General Min-Cost Network Flow problems may serve as justification of applying these algorithms in practice for real problems. The statement of GMNF problem is broad enough, and thus many practical problems may fall under this formulation, which means that these problems may be solved correctly using BP algorithms.

The future research is required to determine the speed of convergence of BP for this Generalized Min-Cost Network Flow, since in our paper we have only proved that after the finite iterations of the algorithms the  answer will not change and be the correct one. However, since $\sigma(x^*)$ in \ref{Th} may be arbitrary small for some problems, the number of iterations until the algorithm will give the correct answer may be arbitrary big, and further analysis is needed to reasonably bound the number of steps that should be done.

\addtolength{\textheight}{-12cm}   





\bibliographystyle{IEEEtran}
\bibliography{main.bib}

\begin{thebibliography}{10}
\providecommand{\url}[1]{#1}
\csname url@samestyle\endcsname
\providecommand{\newblock}{\relax}
\providecommand{\bibinfo}[2]{#2}
\providecommand{\BIBentrySTDinterwordspacing}{\spaceskip=0pt\relax}
\providecommand{\BIBentryALTinterwordstretchfactor}{4}
\providecommand{\BIBentryALTinterwordspacing}{\spaceskip=\fontdimen2\font plus
\BIBentryALTinterwordstretchfactor\fontdimen3\font minus
  \fontdimen4\font\relax}
\providecommand{\BIBforeignlanguage}[2]{{%
\expandafter\ifx\csname l@#1\endcsname\relax
\typeout{** WARNING: IEEEtran.bst: No hyphenation pattern has been}%
\typeout{** loaded for the language `#1'. Using the pattern for}%
\typeout{** the default language instead.}%
\else
\language=\csname l@#1\endcsname
\fi
#2}}
\providecommand{\BIBdecl}{\relax}
\BIBdecl

\bibitem{wainwright2008graphical}
M.~J. Wainwright, M.~I. Jordan \emph{et~al.}, ``Graphical models, exponential
  families, and variational inference,'' \emph{Foundations and
  Trends{\textregistered} in Machine Learning}, vol.~1, no. 1--2, pp. 1--305,
  2008.

\bibitem{murphy1999loopy}
K.~P. Murphy, Y.~Weiss, and M.~I. Jordan, ``Loopy belief propagation for
  approximate inference: An empirical study,'' in \emph{Proceedings of the
  Fifteenth conference on Uncertainty in artificial intelligence}.\hskip 1em
  plus 0.5em minus 0.4em\relax Morgan Kaufmann Publishers Inc., 1999, pp.
  467--475.

\bibitem{pearl1982reverend}
J.~Pearl, \emph{Reverend Bayes on inference engines: A distributed hierarchical
  approach}.\hskip 1em plus 0.5em minus 0.4em\relax Cognitive Systems
  Laboratory, School of Engineering and Applied Science, University of
  California, Los Angeles, 1982.

\bibitem{yedidia2003understanding}
J.~S. Yedidia, W.~T. Freeman, and Y.~Weiss, ``Understanding belief propagation
  and its generalizations,'' \emph{Exploring artificial intelligence in the new
  millennium}, vol.~8, pp. 236--239, 2003.

\bibitem{kolmogorov2005optimality}
V.~Kolmogorov and M.~J. Wainwright, ``On the optimality of tree-reweighted
  max-product message-passing,'' in \emph{Proceedings of the Twenty-First
  Conference on Uncertainty in Artificial Intelligence}.\hskip 1em plus 0.5em
  minus 0.4em\relax AUAI Press, 2005, pp. 316--323.

\bibitem{aji2000generalized}
S.~M. Aji and R.~J. McEliece, ``The generalized distributive law,'' \emph{IEEE
  transactions on Information Theory}, vol.~46, no.~2, pp. 325--343, 2000.

\bibitem{horn1999iterative}
G.~B. Horn, ``Iterative decoding and pseudo-codewords,'' Ph.D. dissertation,
  California Institute of Technology, 1999.

\bibitem{mezard2002analytic}
M.~M{\'e}zard, G.~Parisi, and R.~Zecchina, ``Analytic and algorithmic solution
  of random satisfiability problems,'' \emph{Science}, vol. 297, no. 5582, pp.
  812--815, 2002.

\bibitem{richardson2001capacity}
T.~J. Richardson and R.~L. Urbanke, ``The capacity of low-density parity-check
  codes under message-passing decoding,'' \emph{IEEE Transactions on
  information theory}, vol.~47, no.~2, pp. 599--618, 2001.

\bibitem{weiss2001optimality}
Y.~Weiss and W.~T. Freeman, ``On the optimality of solutions of the max-product
  belief-propagation algorithm in arbitrary graphs,'' \emph{IEEE Transactions
  on Information Theory}, vol.~47, no.~2, pp. 736--744, 2001.

\bibitem{gamarnik2012belief}
D.~Gamarnik, D.~Shah, and Y.~Wei, ``Belief propagation for min-cost network
  flow: Convergence and correctness,'' \emph{Operations Research}, vol.~60,
  no.~2, pp. 410--428, 2012.

\bibitem{birmele2013optimal}
E.~Birmel{\'e}, R.~Ferreira, R.~Grossi, A.~Marino, N.~Pisanti, R.~Rizzi, and
  G.~Sacomoto, ``Optimal listing of cycles and st-paths in undirected graphs,''
  in \emph{Proceedings of the twenty-fourth annual ACM-SIAM symposium on
  Discrete algorithms}.\hskip 1em plus 0.5em minus 0.4em\relax Society for
  Industrial and Applied Mathematics, 2013, pp. 1884--1896.

\bibitem{bayati2008exactness}
M.~Bayati, C.~Borgs, J.~Chayes, and R.~Zecchina, ``On the exactness of the
  cavity method for weighted b-matchings on arbitrary graphs and its relation
  to linear programs,'' \emph{Journal of Statistical Mechanics: Theory and
  Experiment}, vol. 2008, no.~06, p. L06001, 2008.

\bibitem{bayati2008max}
M.~Bayati, D.~Shah, and M.~Sharma, ``Max-product for maximum weight matching:
  Convergence, correctness, and lp duality,'' \emph{IEEE Transactions on
  Information Theory}, vol.~54, no.~3, pp. 1241--1251, 2008.

\bibitem{sanghavi2008linear}
S.~Sanghavi, D.~Malioutov, and A.~S. Willsky, ``Linear programming analysis of
  loopy belief propagation for weighted matching,'' in \emph{Advances in neural
  information processing systems}, 2008, pp. 1273--1280.

\end{thebibliography}

\end{document}